\documentclass[12pt]{article}

\usepackage{arxiv}
\usepackage{caption}
\usepackage[utf8]{inputenc} % allow utf-8 input
\usepackage[T1]{fontenc}    % use 8-bit T1 fonts
\usepackage{hyperref}       % hyperlinks
\usepackage{url}            % simple URL typesetting
\usepackage{booktabs}       % professional-quality tables
\usepackage{amsfonts}       % blackboard math symbols
\usepackage{nicefrac}       % compact symbols for 1/2, etc.
\usepackage{microtype}      % microtypography
\usepackage{graphicx}
\usepackage[numbers]{natbib}
\usepackage{doi}
\usepackage{subcaption} % For side-by-side figures
\usepackage{float}      % For more precise figure placement
\usepackage{tikz}
\usetikzlibrary{arrows.meta,positioning,calc,shapes.geometric,shapes.symbols,shapes.misc,fit,backgrounds}
\usepackage{amsmath}
\tikzset{>=Latex}
\usepackage{rotating}

\title{The Thinking Therapist: Training Large Language Models to Deliver Acceptance and Commitment Therapy using Supervised Fine-Tuning and Odds Ratio Policy Optimization}

\author{
	Talha Tahir \\
	Department of Psychiatry\\
	University of Toronto\\
	\texttt{talhat.tahir@mail.utoronto.ca} \\
}

\date{}

\renewcommand{\headeright}{}
\renewcommand{\undertitle}{}
\renewcommand{\shorttitle}{The Thinking Therapist}

\hypersetup{
hidelinks=true,
pdftitle={The Thinking Therapist},
pdfsubject={cs.AI, cs.CL},
pdfauthor={Talha Tahir},
pdfkeywords={Large Language Models, Psychotherapy, Acceptance and Commitment Therapy, Reinforcement Learning, Chain-of-Thought, ORPO},
}

\begin{document}
\maketitle

\begin{abstract}
	Acceptance and Commitment Therapy (ACT) is a third-wave cognitive behavioral therapy with emerging evidence of efficacy in several psychiatric conditions. This study investigates the impact of post-training methodology and explicit reasoning on the ability of a small open-weight large language model (LLM) to deliver ACT. Using synthetic ACT transcripts generated by \texttt{Mistral-Large}, we trained \texttt{Llama-3.2-3b-Instruct} with two distinct approaches, supervised fine-tuning (SFT) and odds ratio policy optimization (ORPO), each with and without an explicit chain-of-thought (COT) reasoning step. Performance was evaluated by comparing these four post-trained variants against the base Instruct model. These models were benchmarked in simulated therapy sessions, with performance quantitatively assessed on the ACT Fidelity Measure (ACT-FM) and the Therapist Empathy Scale (TES) by an LLM judge that had been fine-tuned on human evaluations. Our findings demonstrate that the ORPO-trained models significantly outperformed both their SFT and Instruct counterparts on ACT fidelity ($\chi^2(5) = 185.15, p < .001$) and therapeutic empathy ($\chi^2(5) = 140.37, p < .001$). The effect of COT was conditional as it provided a significant benefit to SFT models, improving ACT-FM scores by an average of 2.68 points ($p < .001$), while offering no discernible advantage to the superior ORPO or instruct-tuned variants. We posit that the superiority of ORPO stems from its ability to learn the therapeutic `process' over imitating `content,' a key aspect of ACT, while COT acts as a necessary scaffold for models trained only via imitation. This study establishes that preference-aligned policy optimization can effectively instill ACT competencies in small LLMs, and that the utility of explicit reasoning is highly dependent on the underlying training paradigm.
\end{abstract}

\section{Introduction}
Cognitive behavioral therapy (CBT) is a widely employed evidence-based psychotherapy utilized for the treatment of many psychiatric disorders \cite{Hofmann2012, Cuijpers2023}. Variants of CBT are often conceptualized as belonging to one of three ``waves'' \cite{Hayes2006, Hayes2021}. The third-wave is the most recent and represents a shift from symptom reduction to psychological flexibility. Acceptance and Commitment Therapy (ACT) belongs to this third-wave of CBT modalities \cite{Hayes2006, Hayes2021}. ACT has already demonstrated efficacy for treating several conditions, including, depression, obsessive compulsive disorder and chronic pain \cite{French2017, Ma2023, st2014, Feliu-Soler2018}. In simple terms, the goal of ACT is not to eliminate difficult thoughts and feelings but to change the way an individual relates to them \cite{Hayes2006}. 

The principles of ACT have utility even outside the frame of a multi-session therapy course. Interventions such as cognitive defusion or values clarification are effective general-purpose tools for tackling difficult situations and emotions \cite{French2017}. Unfortunately, ACT, like so many other therapy modalities, is difficult to access. Barriers include lack of funds, a paucity of providers and the stigma associated with seeking therapeutic interventions \cite{collins2004gaps}. 

One way in which access to ACT could be increased, is through the utilization of large language models (LLMs). There are many instances of LLMs being deployed in therapy or therapy adjacent contexts with emerging evidence of substantial benefit \cite{Wasenmller2024, Na2024, Kian2024, Iftikhar2024, Tahir2024}. Over the past year, `reasoning models' have become increasingly common \cite{Pan2025}. These models explicitly articulate the intermediate steps that lead to their final output \cite{Ma2023, Wei2022}. The ongoing success of these models in domains such as mathematics and code generation, raises a question as to their efficacy in therapeutic interactions \cite{Liang2024}. 

At the time of writing, to our knowledge there are no publications that evaluate the impact of introducing chain-of-thought or other reasoning approaches on the ability of models to produce therapeutic responses. One might intuitively expect that producing reasoning tokens at inference time may increase the depth and nuance of therapeutic responses by allowing the model to better mentalize the patient. Testing the empirical foundations of this intuition is the focus of our paper. 

Our paper aims to provide an initial foray into investigating the therapeutic performance of `reasoning models' in a simulated environment. We chose ACT as our therapeutic modality as its principles and interventions are transdiagnostic and are useful even outside of a formal therapy course \cite{French2017}. Furthermore, like other third-wave CBT approaches, ACT is quite flexible and practical, allowing us to use its principles in single session interactions without having to spend much time socializing the patient to the ACT philosophy \cite{Hayes2006}. We hypothesized that explicitly modeling the `reasoning' prior to producing the therapeutic response would improve the quality and consistency of LLM delivered ACT interventions. 

To test our hypothesis, we evaluated the therapeutic efficacy of several models developed from a single base model, \texttt{Llama-3.2-3b-Instruct}. These variants were trained using supervised fine-tuning and odds ratio policy optimization (ORPO), both with and without an explicit chain-of-thought (COT) reasoning process \cite{Ouyang2022, hong2024orpo, Wei2022}. To simplify the experimentation process and to better highlight the strengths of each of the approaches, we opted to test the models in one-session interactions with simulated patients rather than modeling an entire course of therapy.

\section{Methods}
\subsection{Overview}
Our study design consisted of three sequential phases. The first phase involved the generation and validation of synthetic ACT transcripts. The second phase entailed using the synthetic data generated in phase 1 to train the \texttt{Llama-3.2-3b-Instruct} model with supervised and reinforcement learning post-training techniques. The last phase focused on testing and evaluating the performance of the trained models through the use of simulated patient interactions. 

\subsection{Synthetic Data Generation} 
We began the synthetic data generation process with the creation of 100 unique patient profiles. Each profile was generated through the random selection of attributes across several categories, including, demographics, clinical presentation, psychosocial factors and interpersonal style.

The patient profiles were then used in the production of 100 synthetic ACT transcripts. Each transcript contained 25 turns of conversation. \texttt{Mistral-Large} was employed to generate both patient and therapist responses. Therapist responses were structured to have the reasonsing tokens appear after the \texttt{<thinking>} tag and final response tokens appear after the \texttt{<answer>} tag. 

For quality assurance, \texttt{DeepSeek-Chat} was used as an LLM supervisor. Patient responses were inspected for consistency and realism whereas therapist responses were inspected for structure, ACT fidelity, context appropriateness and coherence. The LLM supervisor provided a binary `yes' or `no' verdict as to whether revision was required and offered textual feedback for improvement. Responses that required regeneration were sent back to \texttt{Mistral-Large} with the feedback from the LLM supervisor. Employing \texttt{DeepSeek-Chat} as the supervisor provided an orthogonal assessment of quality, to reduce model-specific artifacts and prevent compounding of the generator's inductive biases.

The generated transcripts were then manually reviewed by the author for quality. Each transcript was assessed using the ACT Fidelity Measure (ACT-FM) and the Therapist Empathy Scale (TES) \cite{ONeill2019, Decker2014}. The ACT-FM scale is meant to assess the degree to which a therapist's behavior is consistent with the ACT model whereas the TES is a modality agnostic tool used to evaluate therapeutic empathy \cite{ONeill2019, Decker2014}. The 50 transcripts with the highest composite score on the aforementioned instruments were retained for the training process and the remainder were discarded. 

\subsection{Model Training}
We chose \texttt{Llama-3.2-3b-Instruct} as the base model for all post-training. Training was conducted with the help of the Unsloth library and employed Low-Rank Adaptation (LoRA) \cite{unsloth}. Details on hyperparameters can be found in the GitHub repository for the paper. 

We pursued two different training paradigms, supervised fine-tuning (SFT) and reinforcement learning (RL). SFT was used to train the base model on our 50 curated synthetic ACT transcripts both with and without the COT tokens included. ORPO was used as an alternative technique to train the same base model, again both with and without COT. Group Relative Policy Optimization (GRPO) was considered, with LLM judges providing the reward, however, this idea was eventually discarded due to repeat training failures (secondary to model size limitations and difficulty in creating a smooth reward signal) and computational cost. 

\subsection{Model Testing and Evaluation}
The performance of the trained models was evaluated through interactions with simulated patients. The `patient agent' in the simulations was gpt-5-mini and the `therapist agent' was one of six possible models and configurations, instruct-tuned (as released by Meta), instruct-tuned with a prompt that mandated producing responses in a \texttt{Thinking:/Answer:} format, fine-tuned on the synthetic ACT dataset with and without COT, and ORPO trained again with and without COT. The goal was to be able to compare the performance of these six distinct approaches in delivering ACT-informed responses, thereby isolating the impact of supervised fine-tuning, chain-of-thought reasoning, and reinforcement learning on therapeutic quality.

Each model variant participated in 150 simulated therapy sessions, with every session consisting of 25 turns of dialogue. A unique patient profile was generated for every simulated session through the same process as was used in synthetic data generation. The profiles were held consistent across the six model variants to control for confounding. The `patient agent' was only exposed to the portion of the therapist's response that was produced after the `Answer:' prefix. The thinking tokens produced after the `Thinking:' prefix were saved separately and were not included as part of the conversation history passed to the patient agent. 

To quantitatively evaluate the performance of the models, the transcripts generated during the simulations were assessed by a large language model judge, \texttt{Mistral-Large}, that was fine-tuned on human evaluated synthetic ACT transcripts. Each transcript was rated on the ACT-FM and TES. 

\subsection{Analysis}
\subsubsection{Primary Analysis}
The analysis centered around two primary outcome measures. The ACT-FM total consistency score was calculated as the sum of the four consistency subscales (consistent stance, open, aware, and engaged), with higher scores indicating greater adherence to ACT principles. The TES total empathy mean was computed as the average of all nine TES items. Secondary analyses examined the eight ACT-FM subscales individually.

We employed linear mixed-effects models as part of our primary analytic approach \cite{Oberg2007}. The use of mixed effects was motivated by the non-independence of observations within a given patient profile. As previously mentioned, each model variant was evaluated through simulations with the same set of patient profiles. For every dependent variable, we fitted a model with model type as a fixed effect and patient profile ID as a random intercept. Omnibus tests for the main effect of model type were conducted using a Type~III Wald $\chi^{2}$ test on the fixed-effects coefficients \cite{wald1992sequential}. Models were fitted using Restricted Maximum Likelihood (REML) estimation. Subsequent post-hoc pairwise comparisons between model types were conducted using Wald tests on linear contrasts, with p-values adjusted using the Holm--Sidak method to control familywise error rate while maintaining power comparable to Tukey's HSD \cite{Gourieroux1982, Narkevich2020, Abdi2010}.

Secondary analyses involved computing the Pearson correlation coefficients between all ACT-FM subscales and the total TES empathy score to investigate the relationship between therapeutic consistency and empathy \cite{Sedgwick2012}. 

\subsubsection{Exploratory Subgroup Analysis}
Given that we held the patient profiles as fixed across the different model variants we conducted a series of exploratory subgroup analyses to investigate the degree to which model performance was context dependent. For the analysis, we categorized patient profiles based on four main characteristics derived from their generative attributes. Specifically, we used archetype name (e.g. `Intellectualizer'), psychological mindedness (e.g. `low'), interaction style (e.g. `argumentative/resistant') and primary clinical issue (e.g. `social anxiety'). 

In every patient subgroup, we calculated mean ACT-FM total consistency and TES total empathy scores for all six model variants. To formally test for differences between models within these subgroups, we conducted pairwise comparisons using the non-parametric Wilcoxon signed-rank test, as the scores for each model were paired by patient profile. P-values were adjusted using the Benjamini--Hochberg procedure to control the False Discovery Rate (FDR) \cite{mcknight2010mann, thissen2002quick, taheri2013generalization}. We then computed head-to-head win rates to obtain a direct comparison of model effectiveness. For each of the 150 unique patient profiles, we compared the scores of every possible pair of models. A model was credited with a `win' over another if it achieved a higher score on the relevant metric (ACT-FM or TES). Ties were excluded from this calculation. To determine if the observed win rates were statistically different from chance, we performed an exact sign test for each pairwise comparison against the null hypothesis of a 50\% win rate. The total number of wins was used to rank models on their pairwise performance. To quantify the impact of COT on these results, we subtracted the mean score of the non-COT variant from the mean score of the COT variant within each patient subgroup.

As an exploratory analysis, we fitted additional mixed-effects models including interaction terms between model type and patient archetype to examine whether the effectiveness of different models varied systematically across patient types.

To identify the primary drivers of therapeutic quality, we trained a Random Forest Regressor model to predict ACT-FM and TES scores. To prevent data leakage from repeated measures within patient profiles, we used a 5-fold GroupKFold cross-validation, ensuring that all data from a single patient profile remained within the same fold. The model's out-of-fold predictive performance was evaluated using $R^2$ and Mean Absolute Error (MAE) to establish its credibility before interpreting feature importances. Feature importance was then calculated using permutation importance on the held-out test set for each fold. The final importances represent the average decrease in model performance when a feature's values are randomly shuffled, averaged across all cross-validation folds \cite{Rodriguez-Galiano2015}.

\subsubsection{Data Availability}
All python code used for the primary analysis and exploratory subgroup analyses, along with the synthetic training data and evaluation results, can be accessed through the project's \href{https://github.com/ttahir-git/The_Thinking_Therapist_Final_Code}{\textit{GitHub repository}}. The SFT and ORPO models are available at \href{https://huggingface.co/TTahir}{\textit{HuggingFace/TTahir}}.

\section{Results}
\subsection{Overview} 
Our linear mixed-effects model, detailed in Table \ref{tab:mixed_model}, revealed significant differences in therapeutic performance on both primary outcome measures, ACT-FM total consistency score (Wald $\chi^{2}(5) = 185.15,\ p < .001$) and the TES total empathy mean (Wald $\chi^{2}(5) = 140.37,\ p < .001$). For ACT fidelity, fixed effects (model type) explained 14.8\% of the variance (marginal $R^2$), while the full model including the random effect for patient profile explained 28.1\% (conditional $R^2$). Similarly, for therapeutic empathy, model type explained 11.4\% of the variance (marginal $R^2$), and the full model explained 27.1\% (conditional $R^2$). The random intercept for patient profile ID accounted for a significant portion of the variance in both ACT fidelity (Intraclass Correlation Coefficient [ICC] = 0.16, 95\% CI [0.11, 0.20]) and empathy (ICC = 0.18, 95\% CI [0.14, 0.22]).

\begin{table}[H]
\centering
\caption{\textbf{Results of Linear Mixed-Effects Models for Primary Outcomes.}}
\label{tab:mixed_model}
\resizebox{\textwidth}{!}{%
\begin{tabular}{l r@{ }l r r@{ }l r}
\toprule
& \multicolumn{3}{c}{\textbf{ACT-FM Total}} & \multicolumn{3}{c}{\textbf{TES Mean}} \\
\cmidrule(lr){2-4} \cmidrule(lr){5-7}
\textbf{Fixed Effect} & \multicolumn{2}{c}{\textbf{Estimate [95\% CI]}} & \textbf{p-value} & \multicolumn{2}{c}{\textbf{Estimate [95\% CI]}} & \textbf{p-value} \\
\midrule
Intercept (Instruct COT) & 26.87 & [25.88, 27.86] & $<$ .001 & 5.29 & [5.13, 5.45] & $<$ .001 \\
Instruct (no COT) & -0.62 & [-1.91, 0.68] & .351 & 0.15 & [-0.05, 0.36] & .138 \\
ORPO (COT) & 2.69 & [1.40, 3.98] & $<$ .001 & 0.39 & [0.18, 0.59] & $<$ .001 \\
ORPO (no COT) & 2.61 & [1.32, 3.90] & $<$ .001 & 0.47 & [0.27, 0.68] & $<$ .001 \\
SFT (COT) & -2.08 & [-3.37, -0.78] & .002 & -0.22 & [-0.43, -0.02] & .033 \\
SFT (no COT) & -4.75 & [-6.04, -3.46] & $<$ .001 & -0.58 & [-0.78, -0.37] & $<$ .001 \\
\midrule
\textbf{Random Effect} & \multicolumn{3}{c}{\textbf{Variance (ICC)}} & \multicolumn{3}{c}{\textbf{Variance (ICC)}} \\
\midrule
Patient Profile ID & \multicolumn{3}{c}{6.02 (15.6\%)} & \multicolumn{3}{c}{0.18 (17.8\%)} \\
\bottomrule
\end{tabular}
}
\end{table}

\subsection{Analysis of Therapeutic Quality}
\begin{figure}[htbp]
    \centering
    \includegraphics[width=\textwidth]{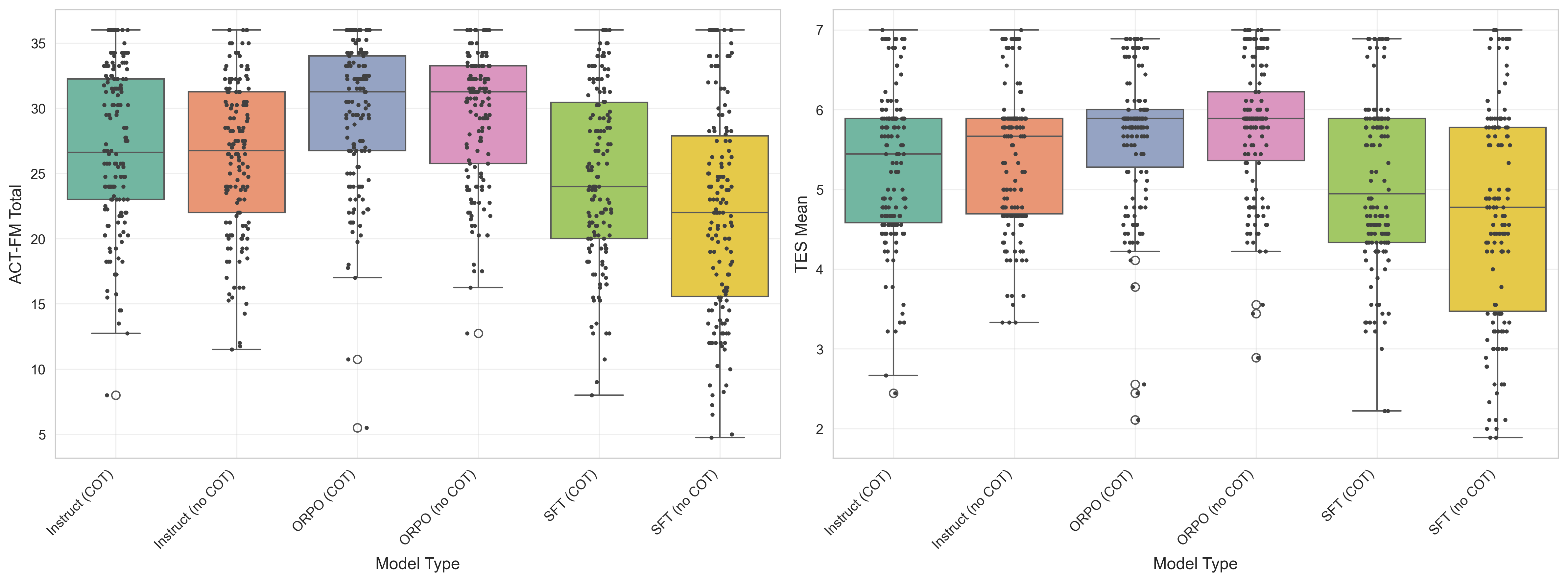}
    \caption{\textbf{Model Performance on Primary Outcomes.} Distribution of primary outcome scores for ACT-FM and TES across the six model variants. The central line indicates the median, the box represents the interquartile range (IQR), and the whiskers extend to 1.5 times the IQR.}
    \label{fig:boxplots}
\end{figure}

A summary of descriptive statistics and the distribution of scores for each model are presented in Table \ref{tab:descriptive_stats} and Figure \ref{fig:boxplots}, respectively. To facilitate interpretation, Table~\ref{tab:emms} reports estimated marginal means from the mixed-effects models for each variant, along with differences versus Instruct (COT), 95\% confidence intervals, standardized effect sizes (Hedges $g$), and Holm-Sidak adjusted $p$-values for both outcomes.

\begin{table}[H]
\centering
\caption{\textbf{Descriptive Statistics for Primary Outcome Measures by Model Type.}}
\label{tab:descriptive_stats}
\begin{tabular}{l c cc}
\toprule
\textbf{Model Variant} & \textbf{N} & \textbf{ACT-FM Total (Mean (SD))} & \textbf{TES Mean (Mean (SD))} \\
\midrule
Instruct (COT) & 150 & 26.87 (6.12) & 5.29 (0.95) \\
Instruct (no COT) & 150 & 26.26 (5.84) & 5.45 (0.90) \\
SFT (COT) & 150 & 24.79 (6.55) & 5.07 (1.03) \\
SFT (no COT) & 150 & 22.12 (8.00) & 4.72 (1.34) \\
ORPO (COT) & 150 & 29.56 (5.46) & 5.68 (0.88) \\
ORPO (no COT) & 150 & 29.48 (5.02) & 5.76 (0.82) \\
\bottomrule
\end{tabular}
\end{table}

\begin{table}[H]
\centering
\caption{\textbf{Estimated marginal means (EMMs) by model with differences vs. Instruct (COT), standardized effects, and Holm-Sidak adjusted $p$-values. Reference = Instruct (COT).}}
\label{tab:emms}
\scriptsize
\begin{tabular}{lrrrrrrrr}
\toprule
& \multicolumn{4}{c}{\textbf{ACT-FM Total}} & \multicolumn{4}{c}{\textbf{TES Mean}} \\
\cmidrule(lr){2-5}\cmidrule(lr){6-9}
\textbf{Model} & \textbf{EMM} & \textbf{$\Delta$ vs Ref [95\% CI]} & \textbf{$g$} & \textbf{$p_{\text{adj}}$} & \textbf{EMM} & \textbf{$\Delta$ vs Ref [95\% CI]} & \textbf{$g$} & \textbf{$p_{\text{adj}}$} \\
\midrule
Instruct (COT)     & 26.87 & ---                         & 0.00  & ---    & 5.29 & ---                         & 0.00  & --- \\
Instruct (no COT)  & 26.26 & $-0.62$ [$-1.91,\,0.68$]   & $-0.11$ & .579  & 5.45 & $+0.15$ [$-0.05,\,0.36$]   & $+0.17$ & .258 \\
ORPO (COT)         & 29.56 & $+2.69$ [$1.40,\,3.98$]    & $+0.47$ & $<.001$ & 5.68 & $+0.39$ [$0.18,\,0.59$]    & $+0.43$ & .002 \\
ORPO (no COT)      & 29.48 & $+2.61$ [$1.32,\,3.90$]    & $+0.46$ & $<.001$ & 5.76 & $+0.47$ [$0.27,\,0.68$]    & $+0.52$ & $<.001$ \\
SFT (COT)          & 24.79 & $-2.08$ [$-3.37,\,-0.78$]  & $-0.36$ & .007 & 5.07 & $-0.22$ [$-0.43,\,-0.02$]  & $-0.25$ & .102 \\
SFT (no COT)       & 22.12 & $-4.75$ [$-6.04,\,-3.46$]  & $-0.83$ & $<.001$ & 4.72 & $-0.58$ [$-0.78,\,-0.37$]  & $-0.64$ & $<.001$ \\
\bottomrule
\end{tabular}

\begin{flushleft}
\footnotesize \textit{Notes.} EMMs from linear mixed-effects models with a random intercept for patient profile. $\Delta$ shows the estimated difference from Instruct (COT) with 95\% CIs from the mixed model. Standardized effect size $g$ uses the model residual SDs: ACT-FM $\sigma_{\varepsilon}\approx 5.71$, TES $\sigma_{\varepsilon}\approx 0.905$. $p_{\text{adj}}$ values are Holm-Sidak adjusted within outcome.
\end{flushleft}
\end{table}

\subsubsection{ACT Fidelity}
ORPO trained models demonstrated greater adherence to ACT principles than their SFT and Instruct counterparts (all Holm--Sidak adjusted p-values $< .001$). Specifically, the ORPO with COT variant (M = 29.56, SD = 5.46) had the highest mean ACT-FM total consistency scores, followed closely by the ORPO non-COT model (M = 29.48, SD = 5.02). There was no significant difference between the two ORPO variants (mean difference = 0.08, adjusted $p = .899$) suggesting that it was the reinforcement learning approach rather than generation of explicit reasoning tokens that contributed most to improvements in ACT fidelity. For context, the standardized mean difference for ORPO (COT) vs. Instruct (COT) on ACT-FM was $\approx0.47$ SD (approximated using the mixed-model residual SD), indicating a moderate effect.

In contrast to the dominance of the ORPO models in ACT adherence, the SFT variants showed the poorest ACT fidelity. The SFT without COT model was the worst overall performer (M = 22.12, SD = 8.00) but did experience a meaningful increase in performance with the addition of COT (mean difference = 2.68, Holm--Sidak--adjusted $p < .001$). SFT with COT scored significantly below Instruct with COT (adjusted $p=.007$) and trended below Instruct without COT (adjusted $p=.078$). SFT without COT was significantly below both Instruct variants (adjusted $p < .001$). The ACT-FM total consistency scores of the COT and non-COT variants of the Instruct models were not significantly different (Instruct with COT: M = 26.87, SD = 6.12; Instruct without COT: M = 26.26, SD = 5.84; mean difference = 0.62, adjusted $p = .579$).

\subsubsection{Therapeutic Empathy}
We observed a similar hierarchy emerge when analyzing therapeutic empathy. The ORPO models led with the highest TES scores (ORPO without COT: M = 5.76, SD = 0.82; ORPO with COT: M = 5.68, SD = 0.88). The differences between the two ORPO variants did not meet the threshold for statistical significance (mean difference = -0.09, adjusted $p = .403$). The repeated absence of significant differences in performance between the two ORPO models in both the domains of ACT fidelity and therapeutic empathy lend additional credence to the idea that ORPO's benefits were not dependent on the generation of explicit reasoning tokens. As a benchmark, the standardized difference for ORPO (no COT) vs. Instruct (COT) on TES was $\approx0.52$ SD (using the model residual SD), a moderate effect.

\begin{figure}[htbp]
    \centering
    \includegraphics[width=\textwidth]{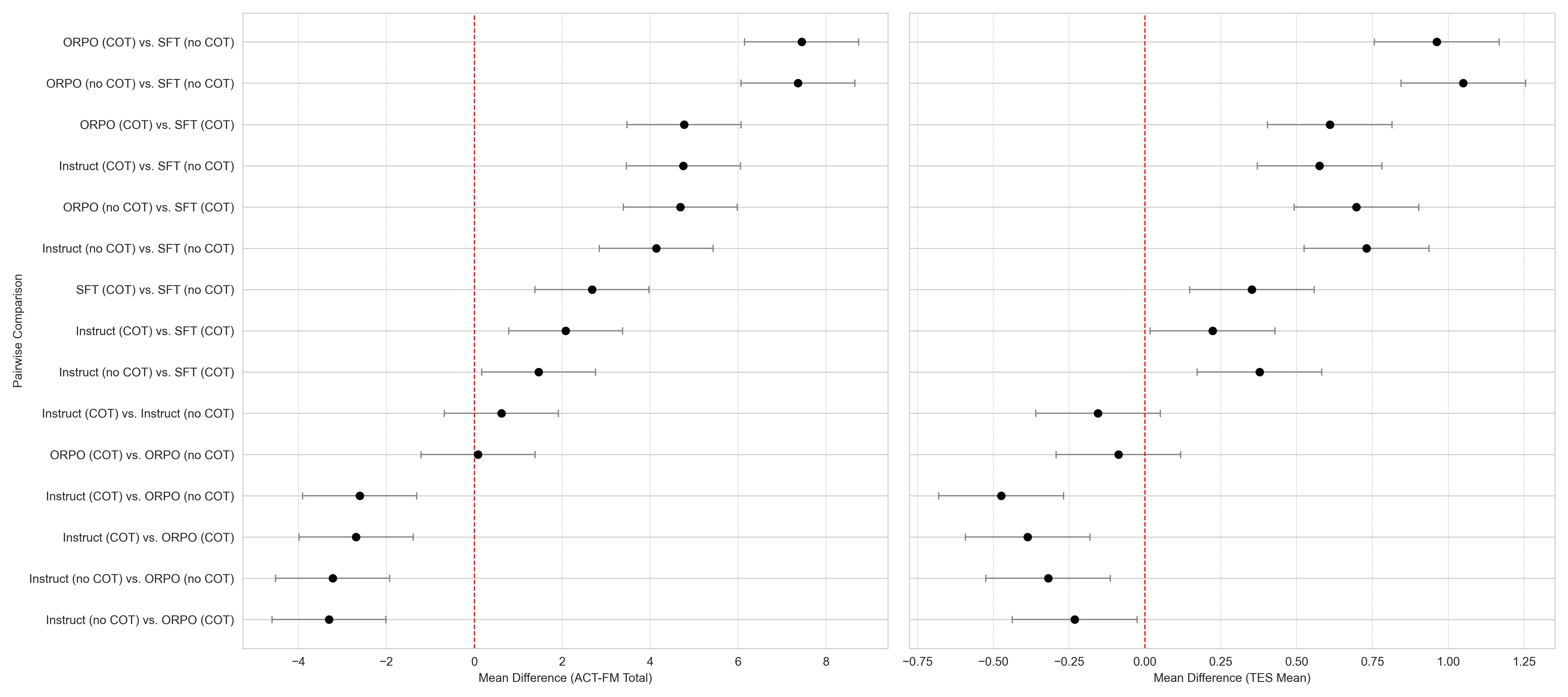}
    \caption{\textbf{Pairwise Model Comparisons on Therapeutic Quality.} Forest plots illustrating estimated mean differences and 95\% confidence intervals from post-hoc pairwise comparisons for ACT-FM and TES. All comparisons are relative to the model listed on the y-axis. A positive difference indicates the model on the x-axis scored higher.}
    \label{fig:forest_plots}
\end{figure}

Similar to our findings in ACT fidelity, the SFT models achieved the lowest empathy scores, particularly the non-COT variant (M = 4.72, SD = 1.34). The addition of COT produced a significant improvement in the empathy of the SFT models (mean difference = 0.35, Holm--Sidak--adjusted $p = .004$), though their performance remained below that of the Instruct and ORPO models. Instruct models showed intermediate levels of empathy, with no significant differences observed between the COT and non-COT variants (adjusted $p = .258$). These pairwise comparisons are visually summarized in Figure \ref{fig:forest_plots}.

\subsection{Relationship Between ACT Fidelity and Empathy} 
A correlation analysis identified strong positive associations between therapeutic empathy and ACT-FM consistency subscales. The TES total empathy score was particularly strongly correlated with consistent stance ($r = .87$), consistent awareness ($r = .84$), and consistent openness ($r = .85$). Meanwhile, negative correlations were observed with all inconsistency subscales ($r$ values ranging from $-.72$ to $-.82$). The strong correlations between empathy and ACT fidelity suggest that they are mutually reinforcing qualities. 

\subsection{Exploratory Analyses}
\subsubsection{Drivers of Therapeutic Quality}
We performed a Random Forest regression using permutation importance to identify the primary drivers of performance. The model demonstrated modest predictive skill on out-of-fold data, achieving an $R^2$ of 0.12 and a Mean Absolute Error (MAE) of 5.03 for ACT-FM scores, and an $R^2$ of 0.08 and an MAE of 0.80 for TES scores. Model type was confirmed as the most influential predictor of both ACT-FM and TES scores, far outweighing any single patient characteristic. For ACT-FM, SFT (no COT) was the most predictive feature (importance = 0.138), followed by SFT (COT) (0.035) and the two ORPO variants (both $\approx0.026$). For TES, SFT (no COT) was again the most important feature (0.099), followed by ORPO (no COT) (0.028) and patient age (0.028).

\subsubsection{Patient Specific Performance}
Our subgroup analyses revealed that the optimal model choice varied based on patient characteristics. For example, we found that for patients with poor psychological mindedness, ORPO without COT performed the best. In contrast, in patients who had moderate psychological mindedness, the ORPO with COT model produced the best ACT-FM and TES scores. Similarly, for the `Intellectualizer' archetype, ORPO with COT was optimal, while for `The Hopeless Skeptic', ORPO without COT was the better choice. These findings were partially supported by an exploratory mixed-effects model which identified a significant interaction between the Instruct non-COT model and the `Hopeless Skeptic' archetype on ACT fidelity ($p = .011$), reinforcing that the model's relative performance is context-dependent. Interestingly, other model-archetype interactions did not reach the threshold for statistical significance. 

\begin{figure}[htbp]
    \centering
    \includegraphics[width=\textwidth]{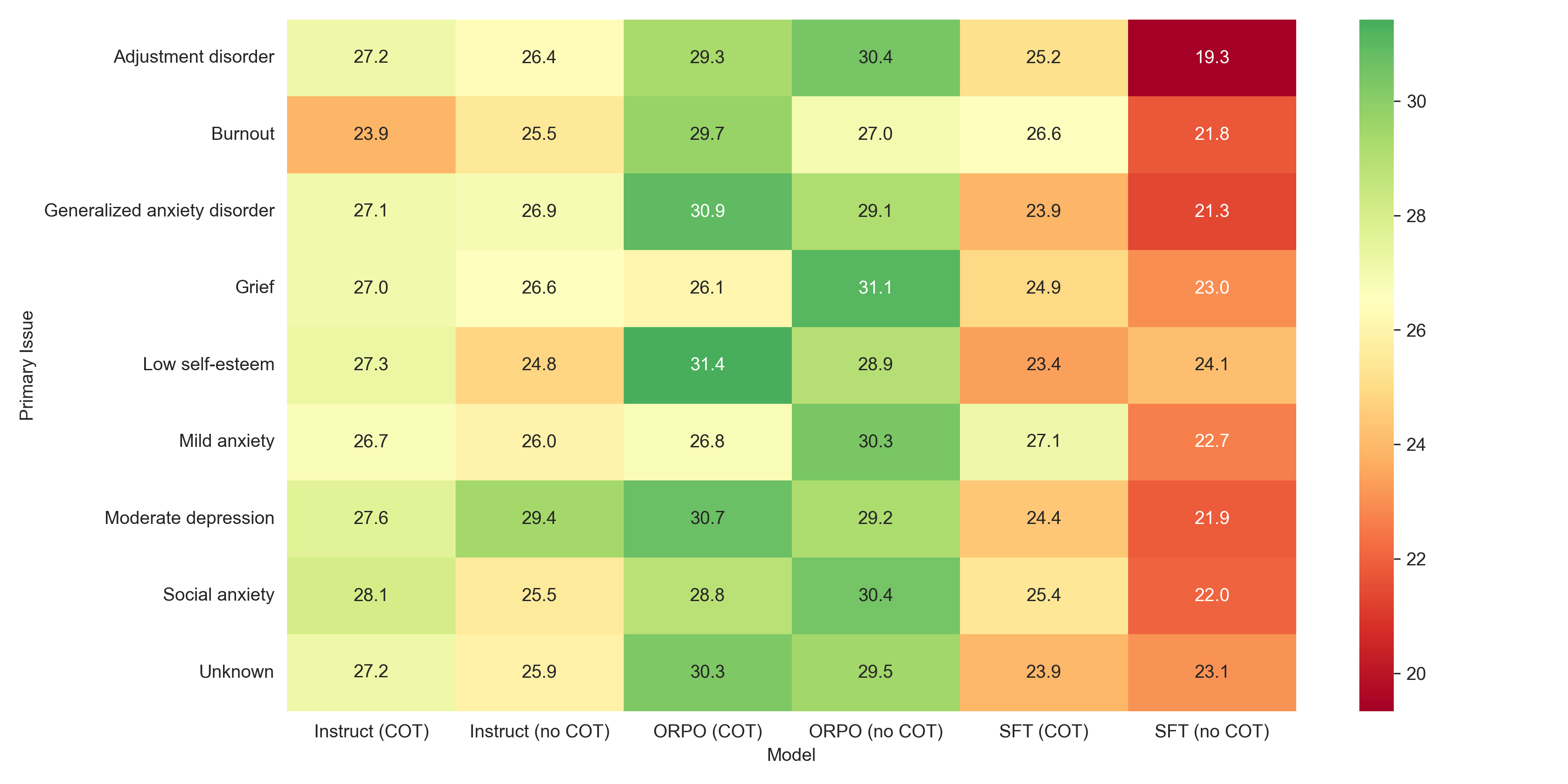}
    \caption{\textbf{Model Performance Across Patient Clinical Presentations.} Heatmap of Mean ACT-FM total consistency scores by model variant and patient primary issue. Deeper green shades indicate better performance, while red shades indicate lower scores.}
    \label{fig:subgroup_heatmap}
\end{figure}

This context dependent performance variability was not limited to psychological mindedness and patient archetypes but extended to the primary clinical issue with which the patient presented. As illustrated in Figure \ref{fig:subgroup_heatmap}, the relative effectiveness of each model variant shifted across clinical presentations such as grief, burnout, and social anxiety. While both ORPO models performed strongly overall, the ORPO (no COT) variant showed a distinct advantage for patients presenting with grief. We note that all subgroup findings should be interpreted as exploratory and descriptive in nature as within subgroup pairwise comparisons employed Benjamini--Hochberg control of FDR and were not familywise-error corrected across all subgroups.

\subsubsection{Head-to-Head Comparisons}
The trend of ORPO superiority emerged again in pairwise comparisons of model variants across patient profiles. In our head-to-head matrices, ORPO with COT defeated SFT without COT in 76.7\% of sessions (95\% CI [69.3\%, 82.7\%], $p < .001$), Instruct with COT in 60.0\% of transcripts (95\% CI [52.0\%, 67.5\%], $p = .006$), and Instruct without COT 69.3\% of the time (95\% CI [61.5\%, 76.2\%], $p < .001$). Additionally, the Instruct COT model won 69.3\% of sessions (95\% CI [61.5\%, 76.2\%], $p < .001$) in its head-to-head comparison with SFT without COT.

\subsubsection{Context Dependent Effects of Chain-of-Thought}
The benefits of producing explicit reasoning tokens varied based on post-training approaches and patient presentations. COT when applied to SFT models provided a consistent and meaningful benefit. Notably, it improved ACT-FM total consistency scores by an average of 2.68 points (Holm--Sidak--adjusted $p < .001$). Contrastingly, for ORPO and Instruct models COT did not lead to appreciable gains in ACT fidelity or therapeutic empathy. For example, in patients with an argumentative or resistant style of engagement, COT improved SFT performance (ACT-FM difference = +3.51) but hindered ORPO performance (ACT-FM difference = $-1.11$). Similarly, for patients presenting with grief, COT had a large negative effect on ORPO model performance (ACT-FM difference = $-5.05$) while still benefitting SFT models. As previously mentioned, these subgroup differences reflect mean shifts within groups and are meant to serve as exploratory signals rather than uniformly significant effects across all contexts. 

\section{Discussion}
\subsection{Principal Findings}
Our study aimed to investigate the impact of producing reasoning tokens on a large language model's ability to deliver ACT in simulated environments. We found that the benefits conferred by this explicit reasoning approach were highly conditional on the post-training methodology. For models trained with SFT, the addition of COT produced a significant increase in both ACT fidelity and therapeutic empathy. However, for both the Instruct and ORPO models, adding COT yielded no improvement in performance. Perhaps the principal finding of our paper is that the most substantial gain in ACT fidelity and empathy came from the preference-aligned policy optimization of ORPO, not from externalized reasoning. Lesser findings of interest include the fact that adherence to ACT principles and therapeutic empathy were strongly correlated, and that patient characteristics greatly influenced model performance as measured by our primary outcomes. 
 
The results of our paper are strengthened by our use of data and models that are small by industry standards. Our dataset in particular, was intentionally quite small, consisting of 50 sets of synthetic transcripts, totaling 1250 discrete prompt-response pairs. That we observed such a clear separation between the ORPO variants and their Instruct and SFT counterparts with our limited sample of data, speaks to both the quality of the training signal and superiority of ORPO as a post-training technique for this task. 

\subsection{Impact of Chain-of-Thought}
Despite the success of COT and other explicit reasoning paradigms in fields such as mathematics and programming, COT did not universally improve therapeutic quality in our study. The one post-training methodology that did benefit from COT was SFT. We hypothesize that in the context of SFT, COT likely functions as a scaffold, compelling the model to deliberate about values, cognitive processes and therapeutic stance prior to producing an answer, thereby improving the structure and consistency of therapeutic interactions. For ORPO, however, such scaffolding is not required as the policy already encodes helpful response patterns optimized against a preference signal. Consequently, forcing the policy to externalize the intermediate steps may make responses more didactic and less attuned. In other words, the benefit of externalized planning may be most apparent in scenarios where the model has only been taught via imitation. 

To answer the question of why explicit reasoning improves performance in mathematics and programming but not psychotherapy, it is helpful to consider the nature of the tasks themselves. Mathematical and programming tasks have an inherent structure that lends itself well to decomposition under hard constraints. In such settings, exposing intermediate steps may assist in search and verification. Therapy on the other hand, is a socio-pragmatic task, in which success depends on turn-by-turn attunement guiding therapeutic intuition on when and how to intervene at the right moment \cite{Rocco2017}. Externalizing a plan can unfavorably skew a policy toward top-down explanation when bottom-up listening and reflection are of greater therapeutic utility. 

Another potential reason for the limited benefits of COT observed in our study is opportunity cost in the context of under-parameterized models. The process of training a 3B parameter model to produce explicit reasoning tokens may consume some of the representational budget needed for attunement to the patient's mental state and subtle adjustments in therapeutic stance. 

\subsection{Superiority of ORPO}
In all of our analyses, the ORPO variants consistently emerged as the best with respect to ACT fidelity and empathy. We feel that the dominance of ORPO is best explained by its unique optimization objective. Unlike SFT, which learns by imitating the content of training examples, ORPO is optimized for the model to learn preferences, in other words, stylistic and relational qualities, that are highly represented in the `winning response' and underrepresented in the `losing response'. Concretely, the odds-ratio objective increases the log-odds of the preferred continuation over a matched alternative, pushing the policy to internalize process-level cues (stance, pacing, reflective listening) rather than merely reproducing surface content. Delivering ACT in particular, can be thought of as a style problem rather than a content one. To use more formal psychotherapy terminology, ACT is a process-focused therapy in which attention is not necessarily placed on the contents of the patient's thoughts, but on their relationship to the process of thinking \cite{Hayes2006, Hayes2021}. By learning this therapeutic `how' instead of a scripted `what,' the ORPO models are not tied to specific examples, making them more robust to context. The small size of our synthetic dataset may also have contributed to the separation in performance between the ORPO models and their SFT and Instruct counterparts. With only 1250 examples, the SFT models likely overfitted to the surface level details or `content', rather than internalizing the nuances of the therapeutic `process'. 

\subsection{Correlations Between Empathy and Adherence}
We observed strong correlations between TES and ACT-FM consistency subscales. While these results have intuitive appeal, two caveats are warranted. First, a single judge model rated both constructs, so common-method variance may inflate the association. Secondly, some inconsistency items inversely mirror consistency items which creates mechanical complementarity that can boost magnitude. Even so, the pattern aligns with clinical intuition that ACT congruent behaviors often feel empathic to recipients \cite{Bricker2011}.

\subsection{Strengths and Limitations}
Our study benefited from a rigorous experimental design. By using a single base model and holding patient profiles constant across 900 simulations, we were able to isolate the effects of SFT, ORPO and COT reasoning. Our use of a mixed-effects analysis further enhanced the causal interpretability of our findings. 

Despite these methodological strengths, the study's reliance on a simulated environment introduces important caveats. While we attempted to create diverse profiles for our simulated patients, the patient agents lacked the lived experience, emotional depth and complex cognitions of the human patients they were meant to model. Similarly, although we attempted to inject human signal into the LLM judge by fine-tuning it on human-evaluated data, it is intuitively unlikely to be able to capture the nuance and contextual understanding possessed by a human rater. The scope of our paper was also limited to single session interactions to simplify experimentation and increase strength of comparisons across simulations. However, psychotherapy is fundamentally a longitudinal process in which outcomes are greatly influenced by the therapeutic alliance \cite{Horvath1993}. Our study did not assess multi-session dynamics such as alliance formation, the navigation of therapeutic ruptures and repairs, or longitudinal changes in symptoms and functioning. 

Our study was also limited by the small size of the training data and base model. It is entirely possible that scaling laws could shift the relative importance of COT and reinforcement learning at larger capacities.

\subsection{Future Directions}
The mixed-effects structure and ICC estimates indicate that patient characteristics have a substantial impact on therapeutic outcomes. The subgroup patterns we observed, such as ORPO without COT performing best for lower psychological mindedness, and ORPO with COT scoring higher for moderate psychological mindedness, raise the intriguing possibility of creating a unified policy through an LLM router \cite{li2025llm}. In practice, a lightweight meta-controller could map a small intake signal (e.g. interaction style, problem focus, psychological mindedness proxy) to a recommended policy variant and COT setting. Dynamically selecting an AI therapist variant through this mechanism would, in some sense, mirror how human clinicians adapt stance and technique to the person in front of them \cite{Rocco2017}. Moreover, relying on routing for creation of an optimal therapy policy would perhaps be more cost and compute efficient than moving to larger models. 

We also posit that model size likely moderates the value of explicit reasoning. Larger models have greater parametric buffer to maintain an internal plan without severely cannibalizing other functions. We hypothesize that COT may positively impact therapeutic outcomes for even RL-tuned policies at higher parameter counts. Similarly, increasing data diversity, should help SFT more than ORPO, narrowing the gap observed in our paper.  

\section{Conclusions}
This study provides an analysis of how explicit reasoning and preference optimization techniques impact the therapeutic quality of AI-delivered ACT. Our results demonstrate that preference-aligned policy optimization (ORPO) can successfully instill ACT competencies in small language models, yielding substantial improvements in both therapeutic fidelity and empathy. The impact of explicit chain-of-thought reasoning was found to be highly conditional, significantly improving the performance of models trained with supervised fine-tuning but offering no discernible benefit to the superior ORPO trained variants. Future studies should focus on exploring the moderating effects of model and data scaling, developing dynamic routing mechanisms to tailor interventions to patient characteristics, refining evaluation methodologies with human raters, and ultimately, pursuing clinical validation.

\section*{Conflicts of Interest} 
The author declares no conflicts of interest that could have influenced the conduct, analysis, or reporting of this research.

\section*{Funding}
This research was independently funded by the primary author (T.T.) without external financial support from any agency, institution, or organization.

\bibliographystyle{unsrtnat}
\bibliography{references}

\end{document}